\documentclass[11pt,english]{article}
\usepackage{times}
\usepackage[T1]{fontenc}
\usepackage{babel}
\usepackage[usenames]{xcolor}
\usepackage{ifpdf}
\ifpdf
\usepackage{pdfpages}
\pdfpageheight\paperheight
\pdfpagewidth\paperwidth
\usepackage{pzccal}
\DeclareFontFamily{OT1}{pzc}{}
\DeclareFontShape{OT1}{pzc}{m}{it}{<-> s * [1.10] pzcmi7t}{}
\DeclareMathAlphabet{\mathpzc}{OT1}{pzc}{m}{it}
\usepackage{graphicx}
\graphicspath{{./figs/}}
\definecolor{linkcolor}{HTML}{4C77EB} 
\definecolor{urlcolor}{HTML}{4C77EB} 
\definecolor{citecolor}{HTML}{4C77EB} 
\definecolor{timecolor}{HTML}{4C77EB} 
\definecolor{draftcolor}{HTML}{4C77EB} 
\definecolor{embargoedcolor}{HTML}{EF4D4E} 
\definecolor{blankcolor}{HTML}{8090AA} 
\definecolor{fontcolor}{HTML}{404040} 
\definecolor{fontcolor}{HTML}{000000} 
\usepackage[pdftex,colorlinks,allcolors=blue]{hyperref}
\hypersetup{allcolors=[rgb]{0.298039215,0.4666666666,0.921568627}} 
\usepackage[pdftex,xcolor]{changebar}
\else
\usepackage[dvips,colorlinks,allcolors=blue]{hyperref}
\hypersetup{allcolors=[rgb]{0.296875,0.46484375,0.91796875}} 
\usepackage[dvips,xcolor]{changebar}
\fi
\usepackage{hyperxmp}
\usepackage[firstpageonly=true]{draftwatermark}
\usepackage[useregional]{datetime2}
\usepackage[hang,small,bf]{caption}
\usepackage{latexsym}
\usepackage{amssymb}
\usepackage[document]{ragged2e}
\RaggedRightParindent=\parindent  
\usepackage{indentfirst} 
\usepackage{lipsum}
\usepackage{calc}
\usepackage{ifthen}
 \newboolean{DRAFT} \setboolean{DRAFT}{false}
 \newboolean{EMBARGOED} \setboolean{EMBARGOED}{false}
 \newboolean{TITLEPAGEABSTRACT}
 \newboolean{CONFIDENTIAL} \setboolean{CONFIDENTIAL}{false}
 \newboolean{FOOTER} \setboolean{FOOTER}{true}
\usepackage{rotating}
\usepackage{fancyhdr}
\usepackage{threeparttable}

\newcommand{\doctitle}{%
A General Measure of Collision Hazard in Traffic}
\newcommand{\docshorttitle}{%
A General Measure of Collision Hazard}
\newcommand{\docauthor}{%
Erik K. Antonsson, Ph.D., P.E., NAE}
\newcommand{\docowner}{%
Streetscope, Inc.}
\ifthenelse{\boolean{CONFIDENTIAL}}%
           {\newcommand{\footertextC}{%
               \href{https://www.streetscope.com}{\docowner}
               {\color{red}Confidential and Proprietary}}}%
           {\newcommand{\footertextC}{%
               \href{https://www.streetscope.com}{\docowner}
           }}

\newcommand{\footerstringC}{\footnotesize\slshape\footertextC}

\fancypagestyle{wicked}{%

\fancyhead[LE]{\normalfont{\thepage}} 
\fancyhead[RO]{\normalfont{\thepage}} 
\fancyfoot[RE,LO]{\mbox{\footnotesize{\docshorttitle}}}
\fancyfoot[CE,CO]{\mbox{\footerstringC}}
\fancyfoot[LE,RO]{\mbox{\footnotesize{\today}}}
\ifthenelse{\boolean{DRAFT}}%
           {\fancyfoot[LE,RO]{\mbox{\footnotesize{
                   \color{black}{\DTMcurrenttime}, {\today}}}}}%
           {}
\ifthenelse{\boolean{DRAFT}}%
           {\fancyfoot[RE,LO]{\mbox{\footnotesize{\color{red}DRAFT}}}}%
           {}
\ifthenelse{\boolean{EMBARGOED}}%
           {\fancyfoot[RE,LO]{\mbox{\footnotesize{\color{blue}EMBARGOED}}}}%
           {}
} 
\fancypagestyle{noheading}{%

\fancyhead[LE]{} 
\fancyhead[RO]{} 
\fancyfoot[RE,LO]{\mbox{\footnotesize{\docshorttitle}}}
\fancyfoot[CE,CO]{\mbox{\footerstringC}}
\fancyfoot[LE,RO]{\mbox{\footnotesize{{\DTMcurrenttime}, {\today}}}}
\ifthenelse{\boolean{DRAFT}}%
           {\fancyfoot[LE,RO]{\mbox{\footnotesize{%
                   \color{black}{\DTMcurrenttime}, {\today}}}}}%
           {}
\ifthenelse{\boolean{DRAFT}}%
           {\fancyfoot[RE,LO]{\mbox{\footnotesize{\color{red}DRAFT}}}}%
           {}
\ifthenelse{\boolean{EMBARGOED}}%
           {\fancyfoot[RE,LO]{\mbox{\footnotesize{\color{blue}EMBARGOED}}}}%
           {}
} 
\fancypagestyle{noheadingfooter}{%

\fancyhead[LE]{} 
\fancyhead[RO]{} 
\fancyfoot[CE,CO]{}
\fancyfoot[LE,RO]{}
\ifthenelse{\boolean{DRAFT}}%
           {\fancyfoot[LE,RO]{\mbox{\footnotesize{%
                   \color{black}{\DTMcurrenttime}, {\today}}}}}%
           {\fancyfoot[LE,RO]{}}
\ifthenelse{\boolean{DRAFT}}%
           {\fancyfoot[RE,LO]{\mbox{\footnotesize{\color{red}DRAFT}}}}%
           {\fancyfoot[RE,LO]{}}
\ifthenelse{\boolean{EMBARGOED}}%
           {\fancyfoot[RE,LO]{\mbox{\footnotesize{\color{blue}EMBARGOED}}}}%
           {\fancyfoot[RE,LO]{}}
} 
\newcommand{\area}{Area}
\newcommand{\subarea}{SubArea}
\newlength{\twospacer}
\newcommand{\rightoddhead}[4]{\fancyhead[RO]{%
\normalfont{\thepage}
\setlength{\unitlength}{0.5in}%
\setlength{\twospacer}{0.90in}%
\begin{picture}(0,0)
\put(1.8,-#1){\makebox(0,0)[b]{\colorbox{#4}{%
\rotatebox[origin=c]{90}{\color{white}{\rule[-0.35in]{0.0mm}{0.45in}%
\shortstack[c]{#2\\#3\\\rule{\twospacer}{0.0mm}}%
}}}}}
\end{picture}
}
}
\newcommand{\leftevenhead}[4]{\fancyhead[LE]{%
\setlength{\unitlength}{0.5in}%
\setlength{\twospacer}{0.90in}%
\begin{picture}(0,0)
\put(-1.74,-#1){\makebox(0,0)[b]{\colorbox{#4}{%
\rotatebox[origin=c]{90}{\color{white}{\rule[0.01in]{0.0mm}{0.45in}%
\shortstack[c]{#2\\#3\\\rule{\twospacer}{0.0mm}}%
}}}}}
\end{picture}
\normalfont{\thepage}
}
}
\makeatletter
\if@twoside%
\newcommand{\blankpage}{%
\mbox{ }\vfill\mbox{ }\hfill
{\color{black}This page intentionally left blank.}
\hfill\mbox{ }\vfill\mbox{ }
\cleardoublepage
}
\else%
\newcommand{\blankpage}{
\pagestyle{empty}
}
\fi
\makeatother
\hypersetup{%
pdftitle={\doctitle},
pdfsubject={Traffic Collision Hazard Measure},
pdfversionid={18},
pdfpubtype={whitepaper},
pdfauthor={{\docauthor}},
pdfauthortitle={Founder and Chief Technology Officer},
pdflang={en},
pdfmetalang={en},
pdfkeywords={Traffic, Collision, Hazard, Measure},
pdfpublisher={\docowner},
pdfcontactaddress={130 W. Union St.},
pdfcontactcity={Pasadena},
pdfcontactpostcode={91103},
pdfcontactcountry={United States},
pdfcopyright={Copyright (c) {\the\year}\xmpcomma\ {\docowner}
  All rights reserved.
  This work is licensed under a
  Creative Commons Attribution-NonCommercial-NoDerivatives 4.0
  International License.
  https://creativecommons.org/licenses/by-nc-nd/4.0/
},
pdflicenseurl={https://www.streetscope.com/},
pdfproducer = {\LaTeX},
pdfcreator  = {\LaTeX},
}
\begin{document}
%
\newcounter{footnoteValueSaver}
\newcounter{pagetemp}
\setcounter{pagetemp}{0}
\newcounter{pagefirst}
\setcounter{pagefirst}{1}
\newcounter{pagelast}
\setcounter{pagelast}{0}
\newcounter{pageoffset}
\setcounter{pageoffset}{\value{page}}
\addtocounter{pageoffset}{-1}
\ifthenelse{\boolean{DRAFT}}{%
\DraftwatermarkOptions{%
  text={DRAFT\\Do Not Distribute},
  scale=0.5
  }
\setboolean{TITLEPAGEABSTRACT}{false}
\begin{titlepage}
  \ifthenelse{\boolean{FOOTER}}{\thispagestyle{noheadingfooter}}{\thispagestyle{empty}}
  \ifthenelse{\boolean{DRAFT}}%
             {\thispagestyle{noheading}%
               \fancyfoot[RE,LO]{\mbox{\footnotesize{\color{red}DRAFTX}}}}%
             {}
  \ifthenelse{\boolean{EMBARGOED}}%
             {\thispagestyle{noheading}%
               \fancyfoot[RE,LO]{\mbox{\footnotesize{\color{blue}EMBARGOED}}}}%
             {}
\noindent
\mbox{%
  \href{https://www.streetscope.com}%
       {\includegraphics*[width=0.25\textwidth]{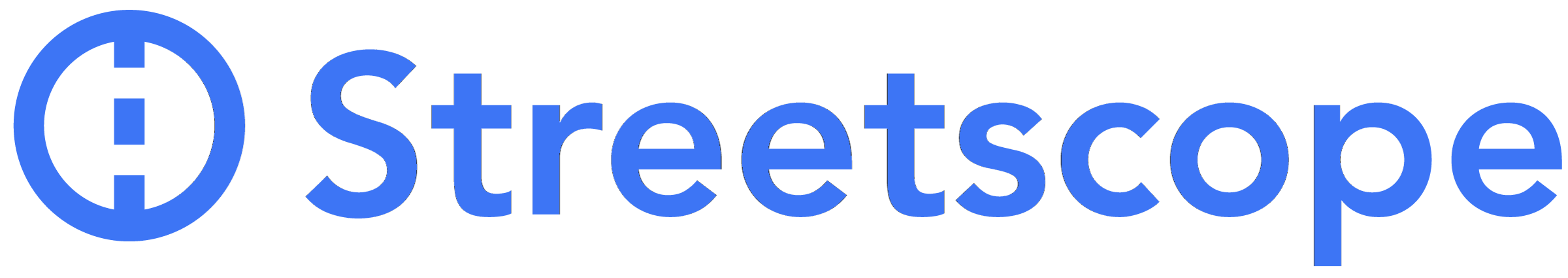}}} \\
\begin{center}
{\Large {\bf {\doctitle}}} \\[12pt]
{\large {\docauthor}} \\[2pt]
\href{https://www.streetscope.com/}{\large {\docowner}} \\
\href{https://www.streetscope.com/}{\footnotesize {https://www.streetscope.com}} \\[12pt]
\ifthenelse{\boolean{DRAFT}}{}{\small{\today}}
\end{center}
%
%
\ifthenelse{\boolean{TITLEPAGEABSTRACT}}{\section*{Abstract}

A collision hazard measure that has the essential characteristics
to provide a measurement of safety that will be useful
to AV developers, traffic infrastructure developers and managers,
regulators and the public is introduced here.
The Streetscope Collision Hazard Measure (SHM\textsuperscript{\texttrademark})
overcomes the limitations of existing measures, and
provides an independent leading indication of safety.

\begin{itemize}
\item Trailing indicators, such as collision statistics, incur
  pain and loss on society, and are not an ethically acceptable
  approach.
\item Near-misses have been shown to be effective predictors of
  the likelihood and severity of incidents.
\item Time-to-Collision (TTC) provides ambiguous indication of
  collision hazards, and requires assumptions about vehicle behavior.
\item Responsibility-Sensitive Safety (RSS), because of its reliance
  on rules for individual circumstances, will not scale up to handle
  the complexities of traffic.
\item Instantaneous Safety Metric (ISM) relies on probabilistic
  predictions of behaviors to categorize events (possible,
  imminent, critical), and does not provide a quantitative measure of
  the severity of the hazard.
\item Inertial Measurement Unit (IMU) acceleration data is not
  correlated with hazard or risk.
\item A new measure,
  based on the concept of near-misses,
  that incorporates both proximity (separation distance)
  and motion (relative speed) is introduced.
\end{itemize}

The new measure presented here gathers movement data about vehicles
continuously and a quantitative score reflecting the hazard encountered
or created is computed nearly continuously, from which the riskiness or
safeness of the behavior of vehicles can be determined.

}{}

\end{titlepage}

\blankpage
}{\DraftwatermarkOptions{stamp=false}}
\ifthenelse{\boolean{EMBARGOED}}{%
\DraftwatermarkOptions{%
  text={Embargoed\\Pre-Release\\Version\\Do Not Distribute},
  scale=0.5
  }
\setboolean{TITLEPAGEABSTRACT}{false}

\blankpage
}{\DraftwatermarkOptions{stamp=false}}
\setboolean{TITLEPAGEABSTRACT}{true}

\newcounter{ntab}
\setcounter{ntab}{1}  
%
\definecolor{tabcolor00}{HTML}{009061} 
\definecolor{tabcolor01}{HTML}{0EBF84} 
\definecolor{tabcolor02}{HTML}{26AEF8} 
\definecolor{tabcolor03}{HTML}{4C77EB} 
\definecolor{tabcolor04}{HTML}{6B59C1} 
\definecolor{tabcolor05}{HTML}{EF4D4E} 
\definecolor{tabcolor06}{HTML}{F27799} 
\definecolor{tabcolor07}{HTML}{FB9933} 
\definecolor{tabcolor08}{HTML}{F3C332} 
\definecolor{tabcolor99}{HTML}{8090AA} 
%
\pagestyle{empty}
\vspace*{12pt}
\noindent
{\footnotesize
The
information
and concepts
presented in this document
are
the property of
{\docowner}
and
are
disclosed in and
covered by one or more published United States and WIPO Patent
Applications, \textit{e.g.},
US~2020/0369270~A1~\cite{EKA:usapp:A};
WO~2020/237207~A1~\cite{EKA:wipo:A}\@.
\vskip12pt
\noindent
\begin{tabular}{rl}
Copyright {\copyright}
{\the\year}
&
\href{https://www.streetscope.com}{\docowner}
\\
All rights reserved.
&
130 W. Union St. \\
&
Pasadena, CA 91103 \\
& U.S.A
\\
& \href{https://www.streetscope.com/}{https://www.streetscope.com}
\\
\end{tabular}
\vskip12pt
\noindent
This work is licensed under a
\href{https://creativecommons.org/licenses/by-nc-nd/4.0/}%
{Creative Commons Attribution-NonCommercial-NoDerivatives 4.0
International License}\@.
\vskip12pt
\noindent
30 29 28 27 26 25 24 23 22 21
({\DTMsetstyle{iso}\DTMnow})
\vskip12pt
} 

\clearpage
\stepcounter{ntab}
\renewcommand{\area}{Table of Contents}
\renewcommand{\subarea}{}
\pagestyle{wicked}
\rightoddhead{\value{ntab}}{\area}{\subarea}{tabcolor00}
\leftevenhead{\value{ntab}}{\area}{\subarea}{tabcolor00}
\fancyhead[LO,RE]{\slshape{\area}} 
\fancyhead[CO,CE]{\slshape{\subarea}} 
\pagenumbering{roman}
\tableofcontents
\listoffigures
\listoftables
\cleardoublepage
\stepcounter{ntab}
\renewcommand{\area}{Introduction}
\renewcommand{\subarea}{Motivation and Objectives}
\pagestyle{wicked}
\rightoddhead{\value{ntab}}{\area}{\subarea}{tabcolor01}
\leftevenhead{\value{ntab}}{\area}{\subarea}{tabcolor01}
\fancyhead[LO,RE]{\slshape{\area}} 
\fancyhead[CO,CE]{\slshape{\subarea}} 
%
\pagenumbering{arabic}
\section{Introduction}
%
\subsection{Motivation and Objectives}

Measurements of traffic safety and risk have been proposed since at least
John Hayward's introduction of Time-to-Collision (TTC) in
1972\@.~\cite{hayward1972_TTC}
Since that time, a variety of additional traffic safety measures have
been proposed and used, including both leading and
trailing indicators.~\cite{wishart2020driving}
This paper introduces a novel leading measure of collision hazards
in traffic that overcomes
difficulties with prior measures, and is quantitative, objective,
continuous,
and general.

Measuring and assessing the safe operation of vehicles in traffic
is essential to the deployment of automated mobility,
as well as improving traffic safety for conventional
human-operated vehicles.
Existing approaches lack key characteristics that will be
required by regulators and the public.

Trailing or lagging indicators of traffic safety are important benchmarks,
however, they are problematical for assessing the risk and safety of
new systems, either roadways, traffic controls, or vehicles.
This is precisely
because they require the accumulation of collision statistics,
and necessarily therefore require collisions to occur.
Traffic collisions incur property damage, injuries, and death and
the ethical problems of inflicting pain and loss on society
(in the process of determining the safety of an AV system)
are unacceptable.

Therefore, for the introduction of new systems, particularly automated
vehicles, leading indicators of traffic risk and safety must be used.
Existing leading road safety indicators have limitations and/or shortcomings,
and a new general measure is needed.
A summary of the most critical shortcomings of several of the most
commonly used measured is presented below.

Importantly, a general quantitative measure of traffic safety and risk
must treat each traffic object (vehicle, pedestrian, bicyclist, \textit{etc.})
as a ``black box'', meaning that only the external behavior of the
object can be utilized in computing the measure.  This is identical to
the conditions of an on-road driving test, where the evaluator simply
observes the actions (behaviors) of the vehicle as it is being driven,
and is not engaged in a dialog with the driver about what he sees nor
what considerations he is making regarding control actions.
The new measure introduced here is such a ``black box'' observational
measure.

A future publication will present
initial results demonstrating correlation of the new measure with
historical collision data, providing validation that
the new measure is useful in predicting likelihood (frequency of occurrence)
and severity of collisions.

\subsection{Existing Methods}
A number of existing approaches to determining traffic safety
are summarized here.
\subsubsection{Near-Misses}

Hayward suggested that near-misses could be an effective leading indicator of
traffic risk and safety:
\begin{quote}
  NEAR-MISS traffic events have been considered for use as predictors
  of accident rate characteristics at roadway locations. The near miss,
  loosely defined, is a traffic event that produces more than an
  ordinary amount of danger to the drivers and passengers involved. Near
  misses would appear to be closely related to the accident pattern
  witnessed at a location and, therefore, could become an attractive
  alternative measure to accident-based safety
  determination.~\cite[page~24]{hayward1972_TTC}
\end{quote}

Other fields have utilized near-misses as leading indicators of risk
for some time~\cite{heinrich1931industrial,busch2021preventing,vanderschaaf2013near},
including civil aviation since at least 1958:
\begin{quote}
  The Aviation Safety Reporting System, or ASRS, is the US Federal
  Aviation Administration's (FAA) voluntary confidential reporting
  system that allows pilots and other aviation professionals to
  confidentially report near misses or close call events in the interest
  of improving aviation
  safety.~\cite{ASRS}
\end{quote}

The chemical processing industry has also implemented near-miss
management systems.
\begin{quote}
  In review of adverse incidents in the [chemical] process industries,
  it is observed, and has become accepted, that for every serious
  accident, a larger number of incidents result in limited impact and an
  even larger number of incidents result in no loss or
  damage.
  \\ $\cdots$ \\
  Despite their limited impact, near misses provide insight into
  accidents that could happen.~\cite[page~445]{phimister2003near}
\end{quote}

``Near-crash'' events were explored in the 2006 DOT/NHTSA
100-Car Naturalistic Driving Study:
\begin{quote}
\begin{quote}
$\bullet$ \underline{Near-Crash}:
Any circumstance that requires a rapid, evasive maneuver by
the subject vehicle, or by any other vehicle, pedestrian, cyclist, or
animal, to avoid a crash. A rapid, evasive maneuver is defined as
steering, braking, accelerating, or any combination of control inputs
that approaches the limits of the vehicle capabilities. As a guide, a
subject vehicle braking greater than 0.5~g or steering input that
results in a lateral acceleration greater than 0.4~g to avoid a crash,
constitutes a rapid maneuver.
\end{quote}
As shown, while these criteria were based somewhat upon quantitative
kinematic criteria, they were subjective in nature. While such
definitions were useful for purposes such as classifying video data,
they were not useful for precisely defining events or as criteria for
other purposes, such as warning algorithms.~\cite[Page~139]{dingus2006100car}
\end{quote}
Despite the limitations of defining and detecting ``near-crash'' events,
the study identified roughly 30 times as many ``near-crashes'' as crash
events.~\cite[Page~141]{dingus2006100car}
The study also identified, particularly in the many scatter plots
of measured data, the serious problems that arise from computations
of traffic characteristics (\textit{e.g.}, TTC and IMU) that are not monotonic:
more severe traffic hazards do not always result in a value that
reflects a greater degree of hazard or risk than less severe hazards.

The majority of near-miss reporting systems rely on observer judgment
as to whether a near-miss occurred or not, and at most have an informal
qualitative assessment of severity.  These approaches are not suitable
for determination of traffic risk and safety because of the wide range
of degree or severity of near-misses in traffic; everything from
inconsequential movement at large distances and low relative speeds to
inches of separation at high speed.  Systems and approaches that
count the occurrence of near misses, as binary events, are not useful
for vehicular traffic.

The collision hazard measure introduced here utilizes the concept of
near-misses among traffic objects (vehicles, pedestrians, bicycles,
stationary objects), and defines a quantitative and continuous
measure of hazard or degree of near-miss.

\stepcounter{ntab}
\renewcommand{\area}{Introduction}
\renewcommand{\subarea}{Existing Methods}
\pagestyle{wicked}
\rightoddhead{\value{ntab}}{\area}{\subarea}{tabcolor02}
\leftevenhead{\value{ntab}}{\area}{\subarea}{tabcolor02}
\fancyhead[LO,RE]{\slshape{\area}} 
\fancyhead[CO,CE]{\slshape{\subarea}} 
\subsubsection{Historical Collision Statistics}

A common practice is to measure the safety of the behavior of
a vehicle on the basis of the frequency of occurrence of collisions.
Typical collision occurrence data is shown in Table~\ref{table:AAA}\@.

\begin{table}[h]
  \caption{Rates of involvement in all police-reported crashes,
    injury crashes, and fatal crashes per 100~million miles driven
    in relation to driver age, United States,
    2014-2015.~\cite{AAA2017crashes}}
  \vspace{3pt}
  \centerline{\begin{tabular}{|c|c|c|c|}
    \hline
    \bf{Age of Driver} & \bf{All Crashes} & \bf{Injury Crashes} & \bf{Fatal Crashes} \\ \hline
    16-17 & 1,432 & 361 & 3.75 \\ \hline
    18-19 &   730 & 197 & 2.47 \\ \hline
    20-24 &   572 & 157 & 2.15 \\ \hline
    25-29 &   526 & 150 & 1.99 \\ \hline
    30-39 &   328 &  92 & 1.20 \\ \hline
    40-49 &   314 &  90 & 1.12 \\ \hline
    50-59 &   315 &  88 & 1.25 \\ \hline
    60-69 &   241 &  67 & 1.04 \\ \hline
    70-79 &   301 &  86 & 1.79 \\ \hline
    80+   &   432 & 131 & 3.85 \\ \hline
  \end{tabular}}
  \label{table:AAA}
\end{table}

Some drivers make less safe decisions (and take less safe actions) than
others, however, the infrequency of collisions, and the many
contributing factors beyond driver decision-making to the occurrence of
collisions, render historical collision statistics of limited use to
evaluate driver performance, or to judge whether a driver (automated or
human) is sufficiently safe to drive, particularly in congested and
complex scenarios.

\subsubsection{Time-to-Collision (TTC)}

\hypertarget{link:TTC}{Some practitioners in the field of traffic safety}
use an estimate of time-to-collision or TTC to indicate whether the vehicle
being analyzed is in a condition of high likelihood of an impending
collision.

Hayward defined time to collision
as follows:
\begin{quote}
  [T]he measure is the time required for two vehicles to collide
  if they continue at their present speeds and
  on the same path.~\cite[page~27]{hayward1972_TTC}
\end{quote}

This measure has significant limitations, particularly
in that the time until a collision will occur is highly dependent on the
speed of the vehicle, the movements of the object with which it might collide,
and the road conditions,
none of which are incorporated into the time-to-collision
measure.\cite{brown2005_adjustedTTC}

Brown noted a particular type of problem with TTC.
\begin{quote}
  Each of the standard crash measures has weaknesses that restrict its
  utility.  Minimum type~I and type~II TTC provide a continuous measure
  of how severe a situation resulted from the driver's response to the
  event so long as the driver does not collide with the other vehicle.
  For this measure, the larger the TTC, the safer the response.  When
  the driver collides, however, the minimum TTC is zero regardless of
  whether the driver barely nudges the other vehicle with a small differential
  velocity or slams into the vehicle with a differential velocity of
  70~mph.~\cite[page~42]{brown2005_adjustedTTC}
\end{quote}  

van~der~Horst identified one of the key limitations of TTC.
\begin{quote}
  [T]he relationship between TTC\textsubscript{\textit{min}} and conflict
  severity scores is not unambiguous; severe conflicts have a low
  TTC\textsubscript{\textit{min}}, but not all conflicts with a low
  TTC\textsubscript{\textit{min}} are regarded as
  severe.~\cite[page~107]{horst2013video}
\end{quote}

A higher-hazard encounter, \textit{e.g.}, one that has the potential to result
in a collision with high relative speed (and therefore would produce
significant damage and/or injury) can have a larger (and therefore
apparently less concerning) TTC than a lower-hazard encounter, such
as one where the potential collision would occur with nearly zero
relative speed.  This example illustrates that TTC is not monotonic
with hazard and risk, as will be discussed further below.

\vspace{2.0pt}

Many attempts to improve TTC have been made, principally focusing on
assumed behavior (\textit{e.g.}, deceleration) of the following vehicle.
\begin{quote}
  Time to collision is an important time-based safety indicator for
  detecting rear-end conflicts in traffic safety evaluations. A major
  weakness of the time to collision notion is the assumption of constant
  velocities during the course of an accident.
  \\ $\cdots$ \\
  Results indicate that in the third case (linear acceleration), the
  average duration of exposure to critical time to collision values is
  greater than the others. So, applying time to collision based on the
  assumption of linear acceleration in collision avoidance systems would
  decrease driver errors more than other
  cases.~\cite[page~294]{saffarzadeh2013generalTTC}
\end{quote}

These limitations render TTC unsuitable as a general measure of collision
risk in traffic.

\subsubsection{Responsibility-Sensitive Safety (RSS)}

\hypertarget{link:RSS}{Responsibility-Sensitive Safety (RSS)}
is a set of five rules intended
to ensure that automated vehicles operate safely.~\cite{shalev2017formal}
These are all sensible rules, however, the full complexity of driving
simply cannot be captured in five rules.\footnote{\href{https://www.mobileye.com/responsibility-sensitive-safety/}{https://www.mobileye.com/responsibility-sensitive-safety/}}
\begin{quote}
\begin{description}
  \item[01. Safe Distance] Enforce a safe following distance from
    a vehicle ahead, based on vehicle speed and stopping ability.
  \item[02. Cutting In] Merge into a lane with sufficient lateral distance
    from other vehicles.
  \item[03. Right of Way]  Give right of way to other vehicles.
  \item[04. Limited Visibility] Be cautious in areas of limited visibility.
  \item[05. Avoid Collisions] [I]f an object suddenly appears in the AV's
    direct path, the AV must avert a crash by veering into the next
    lane, provided it would not cause a different collision.
\end{description}
\end{quote}

While the first two rules include a quantitative measurement of
$d_{\mathit{min}}$, which is defined as the minimum safe distance for those two
maneuvers, RSS does not provide any sort of quantitative measurement of
vehicle safety in traffic.  The first two are behavioral rules that are
to be observed, rather than a quantitative measure of how safely a
vehicle is behaving.  The remaining three rules are important, but
informal and not quantitative, and provide no utility in measuring the
safety of the operation of a vehicle.  Instead, RSS is a set of
guidelines for automated vehicle control decision-making.

Methods that utilize rules for particular
maneuvers or situations, such as RSS,
will not scale well to even a portion of the full range of traffic
scenarios encountered in real traffic, because of the large number of
rules required to accommodate each different type of maneuver.

Rules 3, 4, and 5 assume that vehicle controllers will behave
accordingly, and will have the capability to do so.

A continuous quantitative measure of collision hazards in traffic,
such as the one introduced here,
can assess the risk and safety of a vehicle being operated in
accordance with the RSS rules.

\subsubsection{Instantaneous Safety Metric (ISM)}

\hypertarget{link:ISM}{Instantaneous Safety Metric (ISM)},
developed by the
U.S. National Highway Traffic Safety Administration~\cite{every2017_ISMpaper},
predicts all future positions of one or more vehicles, and examines
the overlap of future reachable regions to determine whether there
is a ``critical'' or ``imminent'' overlap of regions.
An extended approach
(Model Predictive Instantaneous Safety Metric or MPrISM)
is presented in~\cite{weng2020model}\@.

To determine whether a future interaction is ``critical'' or ``imminent''
requires the determination of ``the probability of the driver
choosing to pursue a set of accelerations''~\cite[page~6]{every2017_ISMpaper},
in other words, a prediction of the actions of the operator of each
vehicle is required.

Importantly, the results of computing the future reachable regions and
the possible regions of overlap is one of the four outcomes listed
below.
\begin{quote}
  There are four possible combinations resulting from interaction
  between the possible and unavoidable spaces of two vehicles (Vehicles
  A \& B in this case).~\cite[page~20]{every2017_ISMcharts}
  \begin{enumerate}
  \item The possible space of both vehicles overlap.
    ({\color{red}Possible Interaction})
  \item The unavoidable spaces of both vehicles overlap.
    ({\color{red}Imminent Interaction})
  \item The unavoidable space of Vehicle A overlaps the possible space of
    Vehicle B.
    ({\color{red}Critical interaction for Vehicle A})
  \item The possible space of Vehicle A overlaps the unavoidable space of
    Vehicle B.
    ({\color{red}Critical Interaction for Vehicle B})
  \end{enumerate}
\end{quote}
This is not a quantitative measure of risk of collision; it is
a categorical indicator rather than a metric.  The ISM can determine whether
it is possible for an ``interaction'' to occur in the future, but
other than identifying the future interaction as being either
``possible'', ``imminent'', or ``critical'', the ISM provides no
quantitative indication of the degree of risk nor the severity
of the potential consequences.


\subsection{Disengagements}
\label{sec:disengagements}

\hypertarget{link:disengagements}{The number of disengagements}
of the onboard decision-making system per mile
(where a disengagement is a manual override of the automated
system, such as described in California, California Code of Regulations,
Title ~13, Div.~1, Ch.~1, Article~3.8,
\S227.50\footnote{13~CCR~\S227.50:
  For the purposes of this section, ``disengagement'' means a
  deactivation of the autonomous mode when a failure of the autonomous
  technology is detected or when the safe operation of the vehicle
  requires that the autonomous vehicle test driver disengage the
  autonomous mode and take immediate manual control of the vehicle, or
  in the case of driverless vehicles, when the safety of the vehicle,
  the occupants of the vehicle, or the public requires that the
  autonomous technology be deactivated.  (b) Every manufacturer
  authorized under this article to test autonomous vehicles on public
  roads shall prepare and submit to the department an annual report
  summarizing the information compiled pursuant to subsection~(a) by
  January~1\textsuperscript{st}, of each year.
  })
is also used as an indication
of the performance and safety of the behavior of a vehicle.

The key assumption is that ``disengagements'' occur when hazardous
driving conditions are encountered that the controller does not
handle safely.

The number of ``disengagements'' per mile is not a useful measure
of the behavior or safety or risk of an automated vehicle.
``Disengagements'' can have many causes which may not be related
to the behavior or decision-making system of the vehicle, they
are not repeatable, are subject to the judgment of the safety driver
and therefore occur due to subjective considerations and as
a result are not objective,
and are influenced by the selection of the conditions and scenarios
under which the vehicle is operated and the operational policy or
policies under which the driver operates.~\cite{khattak2020exploratory}

As a result, disengagement rate is not an effective measure of the
hazards encountered by an automated vehicle, and is at best an
indirect indicator of operational risk and safety.

\subsection{Inertial Measurement Unit (IMU)}

\hypertarget{link:IMU}{A number of current approaches to
  measuring risk and safety of vehicles
  utilize data from an onboard inertial measurement unit} (IMU)\@.
This data can indicate rapid deceleration (``hard braking'') or rapidly
executed turns (``swerving''), and it is thought that events of
this type, with accelerations above a threshold, reflect unsafe
operation of the subject vehicle.

While IMU data is readily available, either from onboard accelerometers
or from onboard electronics such as mobile phones, smooth driving
(\textit{i.e.}, with consistently low levels of acceleration)
are at best a poor proxy for the risk or safety of the operation
of a vehicle, for two important reasons:
\begin{enumerate}
\item IMU data is blind to other traffic objects, and therefore
  necessarily takes no account of how the subject vehicle is moving
  in relation to these other traffic objects.  As a result, IMU data
  reflects nothing about how the subject vehicle is interacting with
  traffic.
\item Many cases have demonstrated events where a vehicle was smoothly
  driven into a collision.  Similarly, hard braking occurs when a
  skilled driver avoids a collision, for example in the classic example
  of a child chasing a ball into a lane of traffic from between two parked
  cars.
\end{enumerate}

The use of IMU data relies on the assumption that
rapid decelerations are related to
risky driving behaviors.
The lack of correlation of IMU data with risk and safety of the operation
of a vehicle makes it unsuitable for use as a measure.

%

\clearpage
\stepcounter{ntab}
\renewcommand{\area}{Introduction}
\renewcommand{\subarea}{Nomenclature}
\pagestyle{wicked}
\rightoddhead{\value{ntab}}{\area}{\subarea}{tabcolor03}
\leftevenhead{\value{ntab}}{\area}{\subarea}{tabcolor03}
\fancyhead[LO,RE]{\slshape{\area}} 
\fancyhead[CO,CE]{\slshape{\subarea}} 
\subsection{Definitions}
\begin{description}
\item[Subject Vehicle] is the vehicle whose behavior, safety and risk
  are being analyzed.
\item[Traffic Objects] are other vehicles, pedestrians, bicyclists,
  and other moving objects in a traffic scenario.  In any particular
  traffic scenario there will be $N$ traffic objects (in addition to
  the subject vehicle).  Each traffic object is identified by a number
  from $1$ to $N$\@.
\item[Traffic Scenario] is a physical arrangement of roads and/or streets
  including traffic controls, street markings, curbs, crosswalks,
  and traffic objects.  A typical traffic scenario may last 15-30 seconds.
\item[Near-Miss] is a circumstance where the subject vehicle moves at
  some distance from a traffic object at some speed in relation to the
  traffic object, but a collision does not occur.
\item[Position] $\vec{p}$ is the vector position of the subject vehicle
  ($\vec{p_{v}}$) and each traffic object ($\vec{p_{o_{i}}}$)\@.
\item[Separation Distance] $d_{\mathit{sep}}$ is the nearest distance
  between the subject vehicle and a traffic object, as
  shown in Equation~\ref{eq:dsep}
  and
  illustrated in
  Figures~\ref{fig:basic} and~\ref{fig:converging}\@.
  \begin{equation}
    d_{\mathit{sep}} = \left| (\vec{p_{o_{i}}} - \vec{p_{v}}) \right|
    \label{eq:dsep}
  \end{equation}
  Separation Distance $d_{\mathit{sep}}$ could also be a distance between
  representative points or locations on
  the subject vehicle and a traffic object, such as the center
  point of a bounding box or quadrilateral enclosing or representing
  the vehicle or object.
\item[Separation Distance unit vector] $\vec{u_{d_{\mathit{sep}}}}$
  is the unit vector in the direction
  of $d_{\mathit{sep}}$
  as shown in Equation~\ref{eq:udsep}\@.
\item[Velocity] $\vec{v}$ is the vector velocity of the subject vehicle
  ($\vec{v_{v}}$) and each traffic object ($\vec{v_{o_{i}}}$)\@.
\item[Relative Speed] $S_{\mathit{rel}}$ is the relative scalar speed of the
  subject vehicle in relation to a traffic object, as
  shown in Equation~\ref{eq:converging}
  and
  illustrated
  in Figures~\ref{fig:basic} and~\ref{fig:converging}\@.
  \begin{eqnarray}
    \vec{u_{d_{\mathit{sep}}}} &=& (\vec{p_{o_{i}}} - \vec{p_{v}}) /
    \left| \vec{p_{o_{i}}} - \vec{p_{v}} \right| \label{eq:udsep} \\
    \vec{S_{v}} &=& ( \vec{v_{v}} \cdot (\vec{p_{o_{i}}} - \vec{p_{v}}) ) \times
    \vec{u_{d_{\mathit{sep}}}} \\
    \vec{S_{o_{i}}} &=& ( \vec{v_{o_{i}}} \cdot (\vec{p_{o_{i}}} - \vec{p_{v}}) )
    \times \vec{u_{d_{\mathit{sep}}}}
  \end{eqnarray}
  \begin{eqnarray}
    S_{\mathit{rel}} &=& \left| \vec{S_{v}} - \vec{S_{o_{i}}} \right|
    \label{eq:converging}
    \\
    \mathrm{if}\; S_{\mathit{rel}} &>& 0; \;
    \mathrm{separating/diverging} \\
    \mathrm{if}\; S_{\mathit{rel}} &<& 0; \;
    \mathrm{approaching/converging}
  \end{eqnarray}

  \begin{figure}
    \includegraphics*[width=0.99\textwidth]{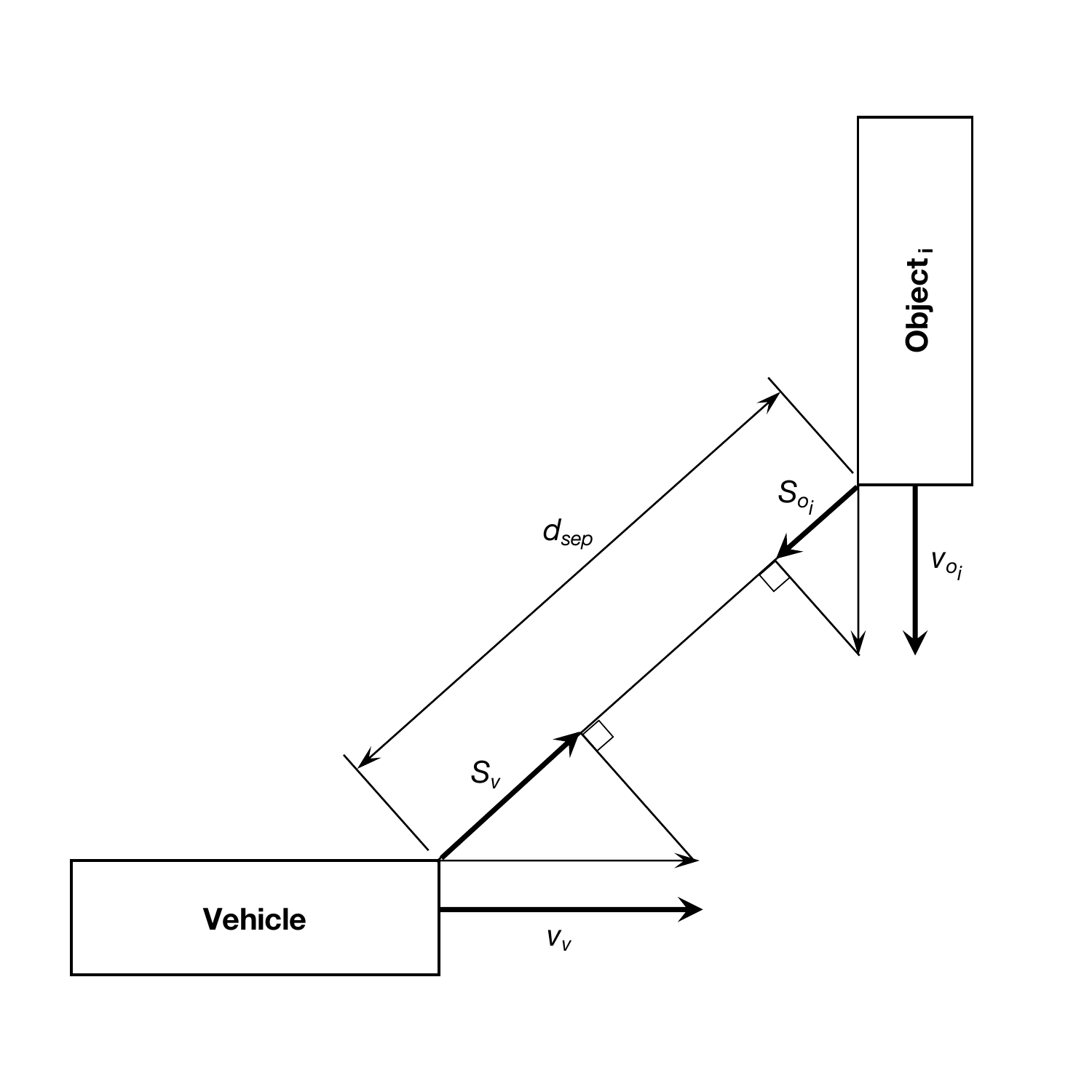}
    \caption{Separation Distance ($d_{\mathit{sep}}$)
      and
      Relative Speed components ($\vec{S_{v}}$ and $\vec{S_{o_{i}}}$)
      between the Subject Vehicle moving with velocity $\vec{v_{v}}$
      and
      Traffic Object\textsubscript{i} moving with velocity $\vec{v_{o_{i}}}$\@.}
    \label{fig:basic}
  \end{figure}

  \begin{figure}
    \includegraphics*[width=0.99\textwidth]{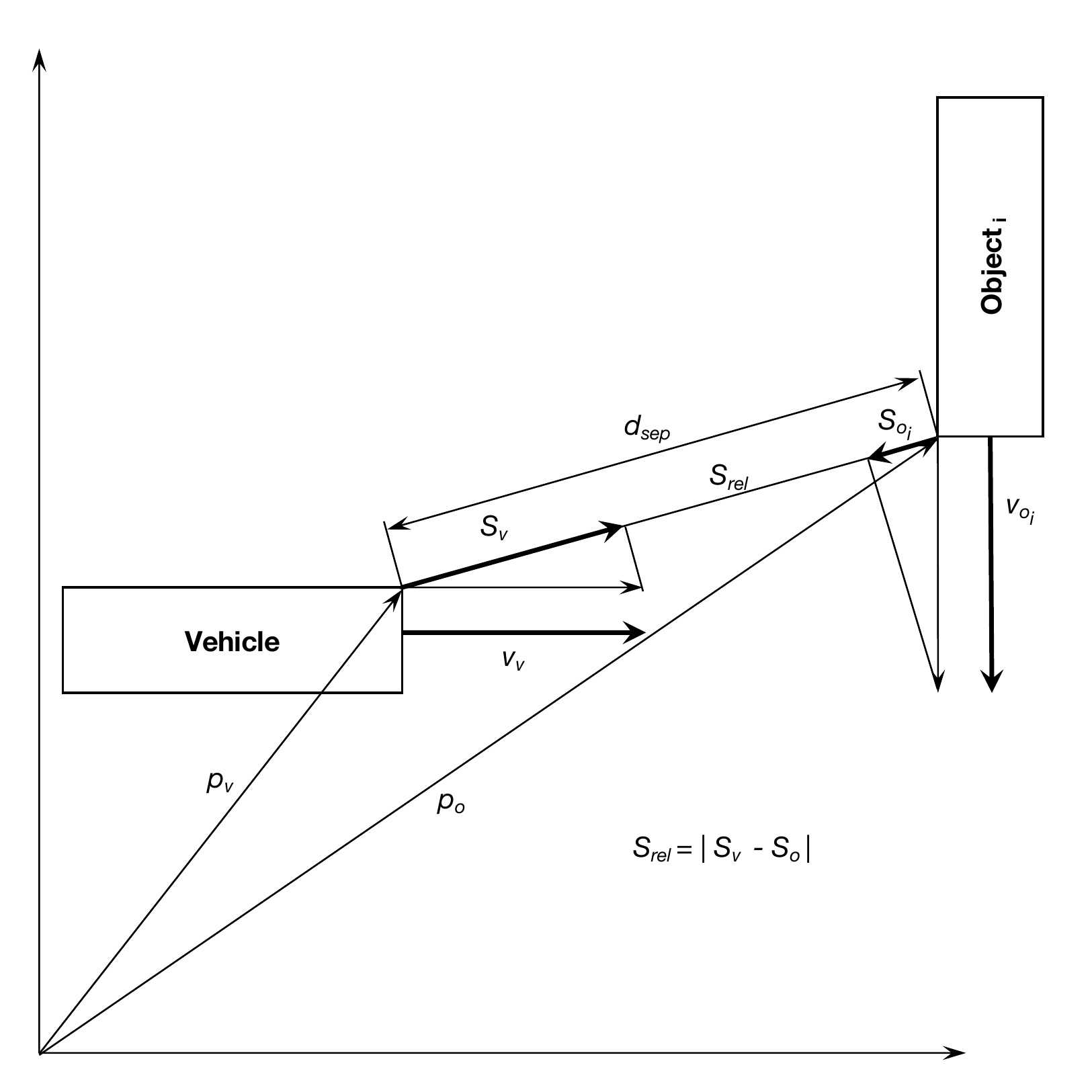}
    \caption{Relative Speed ($S_{\mathit{rel}}$)
      between the Subject Vehicle moving with velocity $\vec{v_{v}}$
      and
      Traffic Object\textsubscript{i} moving with velocity $\vec{v_{o_{i}}}$\@.}
    \label{fig:converging}
  \end{figure}

\item[Grip] is the maximum safe acceleration that the subject
  vehicle can exhibit.  The measure uses both
  $\mathit{braking}\, \mathit{grip}$
  and
  $\mathit{lateral}\, \mathit{grip}$
  to incorporate the different capabilities of
  the subject vehicle to brake and steer.
  Note that $\mathit{grip}$ will be reduced by slippery
  road or street conditions.
\item[Lateral Acceleration] $a_\mathit{lat}$ is the lateral acceleration
  exhibited by the subject vehicle or a traffic object when turning.
\item[Minimum Turning Radius] $r_{t_{\mathit{min}}}$ is the minimum radius of a turn
  that can be executed by the subject vehicle or a traffic object.
  $r_{t_{\mathit{min}}}$ is subject to $\mathit{grip}$ and is an indication
  of the maneuverability of the subject vehicle and each traffic object.
\item[Pair-wise] is the successive consideration of the movement
  of the subject vehicle with each traffic object.
\item[Disengagement] is a manual override of the automated system.
  See section~\ref{sec:disengagements}\@.
\end{description}

\clearpage
\stepcounter{ntab}
\renewcommand{\area}{Introduction}
\renewcommand{\subarea}{Approach}
\pagestyle{wicked}
\rightoddhead{\value{ntab}}{\area}{\subarea}{tabcolor04}
\leftevenhead{\value{ntab}}{\area}{\subarea}{tabcolor04}
\fancyhead[LO,RE]{\slshape{\area}} 
\fancyhead[CO,CE]{\slshape{\subarea}} 
\subsection{Approach}

The novel Streetscope collision hazard measure
(SHM\textsuperscript{\texttrademark})
introduced here
provides a method to quantitatively determine the
varying levels of hazard encountered and created by
the operation of a vehicle, and as a result, estimates of
safety and risk.  Unlike existing assessments of
vehicle safety that pertain to the ability of the subject vehicle to protect
its occupants after a collision begins, this measure assesses the
performance of vehicles prior to and entirely without reference to
collisions.  Collision data is customarily gathered and assessed as
a frequency of occurrence and severity of outcome.
Collisions among existing passenger vehicles
are relatively infrequent (from a statistical standpoint),
limiting the analytic and predictive power of
collision occurrence data.

In contrast, the new measure presented here gathers movement data about
vehicles continuously and a quantitative score reflecting the hazard
encountered or created
(from which the riskiness
or safeness of the behavior of vehicles can be estimated)
is computed nearly continuously,
limited only by the rate at which the sensors provide updated
measurements of the positions
and velocities of vehicles and road conditions.

The new measure is based on the concept of near-misses, rather than
collisions.
The concept of a near-miss comprises two kinematic elements: proximity (near)
and motion (miss)\@.
The new measure incorporates both proximity (separation distance)
and motion (relative speed)\@.

In a near-miss a subject vehicle passes other traffic
objects (\textit{i.e.}, vehicles, pedestrians, bicyclists, any moving or
stationary object in a traffic scenario) but does not collide.  The new
measure is based, in part, on the proximity of the vehicle to each
traffic object and the relative speed between the vehicle and each
traffic object.  In this way, the new measure can assess how near a
vehicle is to a collision with another traffic object, and the relative
speed.  A small distance and low speed can be equivalent to a higher
speed at a larger distance.

Other factors contribute to the degree or severity of a near-miss.
The maneuverability of each traffic object
plays an important role, since traffic objects have limits on
their ability to change speed and direction.
If the road/street surface is slippery, the reduced \textit{grip}
reduces the control authority of each traffic object, increasing
the hazard.  

Therefore, a measure $m$ of
collision hazard between each pair of traffic objects
should be a function
$\mathcal{f}$
of the following four quantities:
\begin{enumerate}
  \item The position $\vec{p}$ of each traffic object
  \item The velocity $\vec{v}$ of each traffic object
  \item The maneuverability $r$ of each traffic object
    (an estimate of the ability to turn and stop)
  \item The \textit{grip} or the coefficient of friction between
    traffic objects and the street/road
  \end{enumerate}
\begin{equation}
  m = \mathcal{f} (\vec{p}, \vec{v}, \mathit{r}, \mathit{grip})
  \label{eq:measure}
\end{equation}

\begin{sidewaysfigure}[ht]
  \includegraphics*[width=1.00\textwidth]{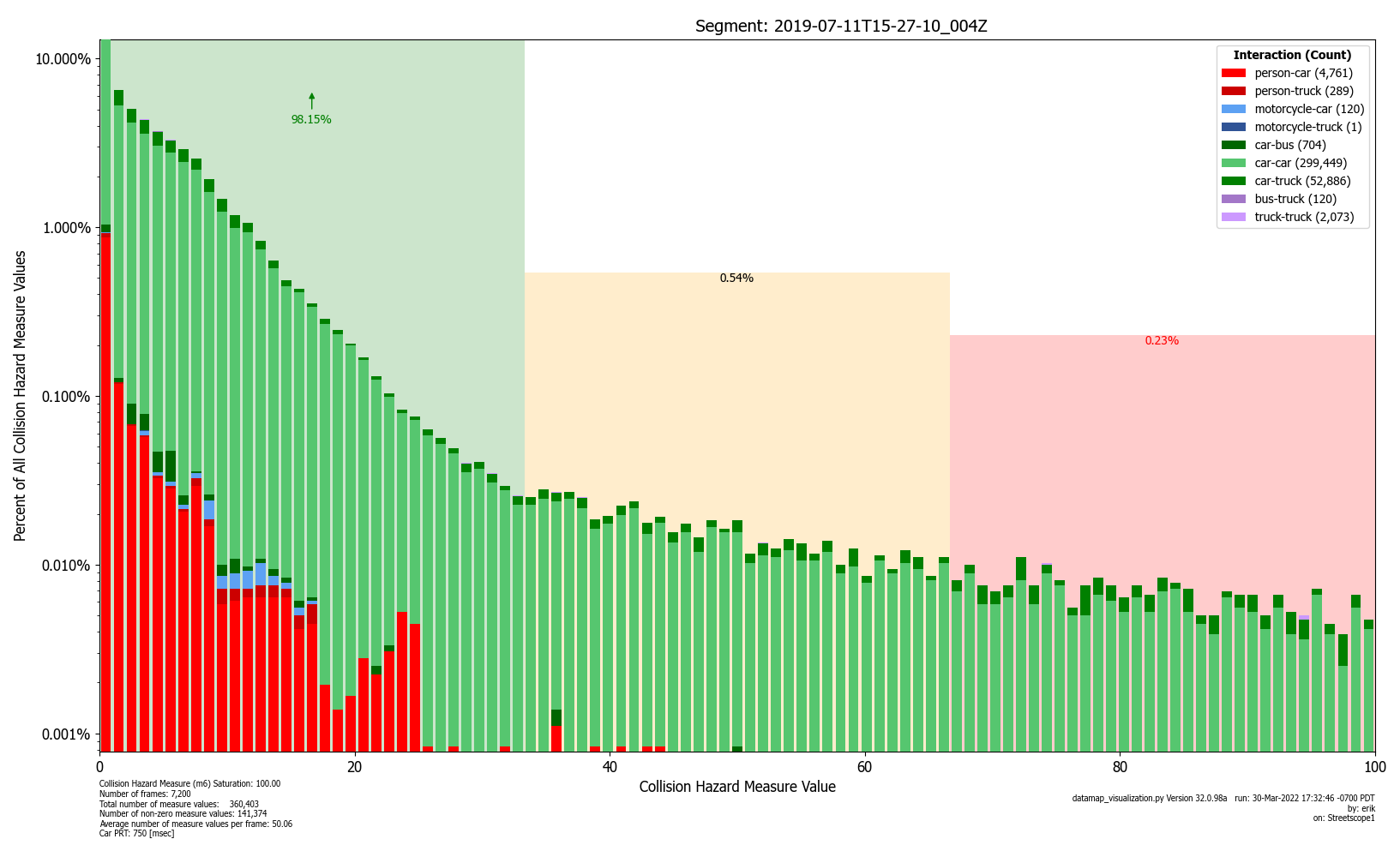}
  \caption{An example histogram of values of the collision hazard measure.}
  \label{fig:histogram}
\end{sidewaysfigure}

Because the new measure is computed nearly continuously for the
subject vehicle and each traffic object,
an aggregation of the data into a score is performed.
One such aggregation is an accumulation of the hazard measure data
into a histogram where the number of times
a hazard value is determined for each value of the hazard value
are accumulated and displayed in a bar-chart.
This results in a chart, similar to the one shown in
Figure~\ref{fig:histogram},
of frequency-of-occurrence vs.\ degree of near-miss.

While a histogram (as with any aggregation of data)
does not fully capture the behavior of
a vehicle in a complex traffic scenario, it represents the key
characteristics that can be compared.

Importantly, for validation purposes, historical collision data
can also be plotted in a histogram, showing the number of times
a collision of a particular severity occurs for each value of
severity.

Experience with near-miss data has shown that, in aggregate, they are
predictive of the likelihood and severity of
incidents.~\cite{vanderschaaf2013near,phimister2003near,muermann2002near,oktem2013improve,kim2017study}

The new measure can be used for evaluation of the performance of an
automated vehicle, in particular the performance of the sensors on
the vehicle, the configuration of the sensors on the vehicle, and the
decision-making performed by the vehicle.
While the measure is intended to be applied to automated vehicles,
it can equally be used to provide a score for human drivers
(and, in-effect, their sensors and decision-making).

The measure can be computed from vehicle and traffic object data
(positions and velocities) that are generated by sensors onboard
one or more vehicles, either moving or stationary, or generated
by stationary sensors, such as video cameras used for traffic
management.

No such driver behavior score or measure exists today,
other than the trailing/lagging indicator: frequency of
occurrence of collisions, and the indicators such as those derived
from IMU data briefly reviewed above.

\clearpage
\stepcounter{ntab}
\renewcommand{\area}{Method}
\renewcommand{\subarea}{}
\pagestyle{wicked}
\rightoddhead{\value{ntab}}{\area}{\subarea}{tabcolor05}
\leftevenhead{\value{ntab}}{\area}{\subarea}{tabcolor05}
\fancyhead[LO,RE]{\slshape{\area}} 
\fancyhead[CO,CE]{\slshape{\subarea}} 
\section{Method}
\subsection{Characteristics}

The key \hypertarget{link:characteristics}{characteristics} of
any effective
collision hazard measure
include:
\begin{enumerate}
\item \hypertarget{link:leading}{\textbf{Leading:}}
    \label{list:leading}
    The measure determines hazards prior to the occurrence of collisions,
    in contrast to collision statistics that determine hazards after
    a number of collisions have occurred.
  \item \hypertarget{link:quantitative}{\textbf{Quantitative:}}
    \label{list:quantitative}
    The result of the computation of the measure is
    a numerical representation of the collision hazard encountered.
  \item \hypertarget{link:continuous}{\textbf{Continuous:}}
    \label{list:continuous}
    The quantitative result is on a continuous scale, \textit{e.g.}, from
    0 (safe) to 100 (a few centimeters from collision), and is
    nearly continuous in time, subject only to the update rate of
    the traffic sensor(s) used.
  \item \hypertarget{link:independent}{\textbf{Independent:}}
    \label{list:independent}
    The computation of the measure relies only on
    external observation of vehicles and traffic objects, and road/street
    surface conditions, and not on sensor data or decision-making
    involved in the control of the vehicle.  The measure considers
    each traffic object to be a ``black box'', not subject to
    internal scrutiny.
  \item \hypertarget{link:direct}{\textbf{Direct:}}
    \label{list:direct}
    The measure directly determines the hazard between traffic objects,
    rather than indirect or proxies for hazard.
  \item \hypertarget{link:repeatable}{\textbf{Repeatable:}}
    \label{list:repeatable}
    Observation of the same behavior in the same
    traffic scenario will produce the same quantitative value of the
    collision hazard measure.
  \item \hypertarget{link:noassumptions}{\textbf{No Assumptions:}}
    \label{list:noassumptions}
    The measure does not make use of assumptions or predictions
    about the actions or behaviors of traffic objects.
  \item \hypertarget{link:monotonic}{\textbf{Monotonic:}}
    \label{list:monotonic}
    A more severe collision hazard will always result in
    a larger value of the collision hazard measure.
    Two identical collision hazards, generated by different traffic conditions,
    will always result in the same value of the collision hazard measure.
    To be at all useful, the measure should at least
    conform with the properties of an
    \textit{Ordinal Scale}, preferably a
    \textit{Ratio Scale}~\cite{%
      stevens1946measurement,%
      britannicaMeasurementScale}
    summarized in Appendix~\ref{app:measurement_scales}
    on Page~\pageref{app:measurement_scales}\@.
  \item \hypertarget{link:objective}{\textbf{Objective:}}
    \label{list:objective}
    No qualitative or subjective input is included
    in the computation of the result.  The result depends solely on
    the measured kinematics of the vehicles and traffic objects,
    and road/street surface conditions.
  \item \hypertarget{link:computable}{\textbf{Computable:}}
    \label{list:computable}
    Given the kinematics (positions and velocities)
    of the subject vehicle and other traffic objects in a scenario,
    and estimates of the capabilities of the subject vehicle to stop and turn,
    the measure can be automatically calculated
    by a machine such as a computer.
  \item \hypertarget{link:scalable}{\textbf{Scalable:}}
    \label{list:scalable}
    The computation of the measure does not depend
    on the complexity or other characteristics of the traffic scenario,
    and therefore naturally and easily scales up to the full range of
    situations and scenarios encountered in real traffic.
    Measures that comprise individual rules or computations for
    specific traffic scenarios are inherently not scalable.
\end{enumerate}

\subsection{Streetscope Collision Hazard Measure (SHM)}

\hypertarget{link:SHM}{The novel measure of vehicle and
  traffic risk and safety} described here
utilizes the position and velocity of the subject vehicle, the
position and velocity of each traffic object,
the road conditions,
and an estimate of the maneuverability of the subject vehicle
and traffic objects
(maximum safe braking deceleration rate and maximum safe turning rate)\@.
Table~\ref{tab:characteristics} presents
a comparison of the characteristics of
selected traffic collision hazard measures.

\begin{table}
  \centering
  \caption{Characteristics of Selected Traffic Collision Hazard Measures}
  \label{tab:characteristics}
  \begin{threeparttable}
    \centering
    \begin{tabular}{|l|c|c|c|c|c|c|}
    \hline
    \hyperlink{link:characteristics}{Characteristic} &
    \hyperlink{link:SHM}{SHM} & 
    \hyperlink{link:TTC}{TTC} &
    \hyperlink{link:RSS}{RSS} &
    \hyperlink{link:ISM}{ISM} &
    \hyperlink{link:disengagements}{Disengagements} &
    \hyperlink{link:IMU}{IMU} \\
    \hline
    \hline
    \ref{list:leading}. \hyperlink{link:leading}{Leading} &
    \checkmark & \checkmark & \checkmark & \checkmark &   & \\
    \hline
    \ref{list:quantitative}. \hyperlink{link:quantitative}{Quantitative} &
    \checkmark & \checkmark & binary & binary & binary  & \checkmark \\
    \hline
    \ref{list:continuous}. \hyperlink{link:continuous}{Continuous} &
    \checkmark & \checkmark & count\tnote{a} & count\tnote{a} & count\tnote{a} & count\tnote{a} \\
    \hline
    \ref{list:independent}. \hyperlink{link:independent}{Independent} &
    \checkmark & \checkmark &   & \checkmark &   & \checkmark \\
    \hline
    \ref{list:direct}. \hyperlink{link:direct}{Direct} &
    \checkmark & \checkmark &   & \checkmark &   &   \\
    \hline
    \ref{list:repeatable}. \hyperlink{link:repeatable}{Repeatable} &
    \checkmark & \checkmark &   & \checkmark &   & \checkmark \\
    \hline
    \ref{list:noassumptions}. \hyperlink{link:noassumptions}{No Assumptions} &
    \checkmark &  &   &   &   &   \\
    \hline
    \ref{list:monotonic}. \hyperlink{link:monotonic}{Monotonic} &
    \checkmark &   &   &   &   &   \\
    \hline
    \ref{list:objective}. \hyperlink{link:objective}{Objective} &
    \checkmark & \checkmark &   & \checkmark &   &   \\
    \hline
    \ref{list:computable}. \hyperlink{link:computable}{Computable} &
    \checkmark & \checkmark & partial\tnote{b}  & \checkmark &   & \checkmark \\
    \hline
    \ref{list:scalable}. \hyperlink{link:scalable}{Scalable} &
    \checkmark & \checkmark &   &   &   & \checkmark \\
    \hline
    \end{tabular}
    \begin{tablenotes}
    \item[a] \footnotesize{Occurrences of events are counted,
      and therefore are not a continuous measure.}
    \item[b] \footnotesize{Two of the RSS rules are computable;
      the remainder are not.}
    \end{tablenotes}
  \end{threeparttable}
\end{table}

In all cases, the measure is computed sequentially for
the subject vehicle in relation to each traffic object.
For a subject vehicle and eight traffic objects, the measure
will be computed in a pair-wise manner:
eight times at each time step, once for
each traffic object in relation to the subject vehicle.

The essence the measure
incorporates the square of the
the relative speed between the subject vehicle and a traffic object
($S_{\mathit{rel}}$)
divided by the distance that separates the vehicle and the object
($d_{\mathit{sep}}$)\@.
\begin{equation}
  m_2 = S_{\mathit{rel}}^{2} / d_{\mathit{sep}}
  \label{m2}
\end{equation}
$m_3$ has the units of $\left[ \mathrm{length}/\mathrm{time}^{2} \right]$
or
$\left[ \mathrm{acceleration} \right]$\@.

This measure has the essential character of near-misses described
above, combining proximity (separation distance) and motion (relative speed)
for each pair of traffic objects in a traffic scenario
(\textit{e.g.},
car\textsubscript{\textit{i}}-pedestrian\textsubscript{\textit{j}})
at each frame of sensor data.
In this case, relative speed is in the numerator, so the measure will
be larger for larger values of relative speed; separation distance is
in the denominator so that the measure will be larger for \emph{smaller}
separation distances.

This matches our perception of near-misses.  A vehicle that
is moving at 0.5~m/s (1~mph) past a pedestrian at a distance of 1~meter
(39~inches) would not be alarming or considered to be particularly dangerous.
In contrast a vehicle that is moving at 30~m/s (67~mph) past a pedestrian
at the same distance would be highly alarming and would be considered to
be seriously dangerous.  Both relative speed and separation distance are
essential characteristics of a quantitative measure of near-misses, and
the hazard that they produce.

The influence of the speed of the subject vehicle in relation to the
traffic object is considerably magnified in $m_2$ compared to
other approaches to quantify near-misses.
This magnification is desired, and is an important
characteristic of the measure since the square of the speed is directly
proportional to the kinetic energy of the subject vehicle in relation to
the traffic object and the dissipation of kinetic energy in a collision
is the cause of damage and injury.

An augmented version of the measure
incorporates the square of the
maximum of the absolute speed of the subject vehicle
($S_{\mathit{abs}}$)
and the relative speed between the subject vehicle and a traffic object
($S_{\mathit{rel}}$)
divided by the distance that separates the vehicle and the object
($d_{\mathit{sep}}$)\@.
\begin{equation}
  m_3 = \mathrm{max}(S_{\mathit{abs}}, S_{\mathit{rel}})^{2} / d_{\mathit{sep}}
  \label{m3}
\end{equation}
$m_3$ also has the units of $\left[ \mathrm{length}/\mathrm{time}^{2} \right]$
or
$\left[ \mathrm{acceleration} \right]$\@.

This approach to combining a compensated (\textit{i.e.}, squared) value of
relative speed with separation distance has the essential characteristic
of monotonicity: a less severe traffic hazard will result in a lower
numerical value of the measure than a more severe traffic hazard.

Importantly, the determination of the measure values makes no
assumptions nor predictions about the behavior or actions of traffic objects;
it simply assesses the current state of the near-miss interaction
between each pair of objects, and does not predict future actions,
decisions or trajectories.

Because the measure can be computed from simple low-cost sensors
that are independent of the automated on-board sensing and decision-making,
the SHM treats the vehicle, and those around it in traffic, as a
``black-box'', and the results are a fully independent measure of
the hazards encountered by a vehicle in traffic, and its responses to
those hazards.

\subsection{Further Augmented Streetscope Collision Hazard Measure}

Additional features of the SHM have been implemented,
including incorporation of road/street surface characteristics
(traction or \textit{grip}\footnote{%
  Road/street traction or \textit{grip} can be estimated from video imagery,
  or can be measured independently periodically, or can be determined from
  an on-board traction-control system.
  Estimates of \textit{grip} will be presented in future publications.})
and estimates of the limits of maneuverability
of each traffic object (\textit{e.g.}, minimum turning radius, braking distance,
perception-reaction time)\@.
These further augmentations will be presented in future publications.

Both U.S. and International patents are pending.~\cite{EKA:usapp:A,EKA:wipo:A}

\clearpage
\stepcounter{ntab}
\renewcommand{\area}{Example}
\renewcommand{\subarea}{Illustrations}
\pagestyle{wicked}
\rightoddhead{\value{ntab}}{\area}{\subarea}{tabcolor06}
\leftevenhead{\value{ntab}}{\area}{\subarea}{tabcolor06}
\fancyhead[LO,RE]{\slshape{\area}} 
\fancyhead[CO,CE]{\slshape{\subarea}} 
\section{Example Illustrations}
\subsection{Simulation}

Simulations were used as the starting point to demonstrate
how the SHM works, and to illustrate its
effectiveness and utility.

Figure~\ref{fig:sim} shows three frames from a simulation of traffic,
overlaid with various displays of the results of the computation of
the SHM\@.
The display shows an overhead view of a 4-way intersection with
various vehicles and other traffic objects.

In the simulation, the subject vehicle (or ego vehicle) is
a car, shown as a bright blue rectangle, traveling in the Eastward direction.
The ego vehicle has a red border in these three frames, illustrating
that it is encountering hazard values that are in the Unsafe range.

There are three other cars in the scenario: a green car heading South,
a light-purple car heading West, and a light blue car traveling East
somewhat ahead of the ego vehicle.

In addition to the cars, there is a bus, shown as the bright purple
elongated rectangle, heading Northward across the intersection.
There are also 3 bicycles, shown as green cigar-shapes;
two pedestrians heading North across the intersection shown as small circles,
and two fire hydrants, shown as small squares.

Since this simulation is an ego-centric scenario, where our interest is
solely in the hazard encountered by the ego vehicle, and not on the
interactions between the other traffic objects, a line is drawn on
the simulation between the ego vehicle and the center of each of
the traffic objects.  These lines illustrate the results of the
computation of the SHM by the width and color
of the line.\footnote{Note that two thresholds have been established
  for values of the SHM: a transition from Safe
  to Hazardous, and a transition from Hazardous to Unsafe.  These
  thresholds, in practice, will be set by regulatory authorities
  to help identify the safety/riskiness of driving behaviors.
  For the purposes of the simulation shown here, the thresholds
  have been set arbitrarily, as an illustration.}
\begin{itemize}
\item A thin white line indicates that there is no hazard being
  encountered between the ego vehicle and the traffic object.  This
  situation occurs, for example, if the ego vehicle and the traffic object
  are moving away from each other.
\item A slightly thicker green line indicates that there is a hazard
  value that has been computed between the ego vehicle and the traffic
  object, but the value is in the Safe range.
\item A thicker yellow line indicates that the collision hazard value
  between the ego vehicle and the traffic object
  is in the Hazardous range.
\item A thick red line indicates that the collision hazard value
  between the ego vehicle and the traffic object
  is in the Unsafe range.
\end{itemize}

The speed of the ego vehicle is displayed in its rectangle,
along with the largest collision hazard value encountered among
the visible traffic objects at this time.
Each traffic object also has its speed and individual collision hazard value
(if visible to the ego vehicle) displayed.

No collisions occur in this simulation, however, there are a number of
very close near-misses, by design, to exercise and demonstrate the
SHM\@.

\newlength{\figheight}
\setlength{\figheight}{0.30\textheight}
\begin{figure}[ht]
  \centerline{\fbox{\includegraphics*[height=\figheight]{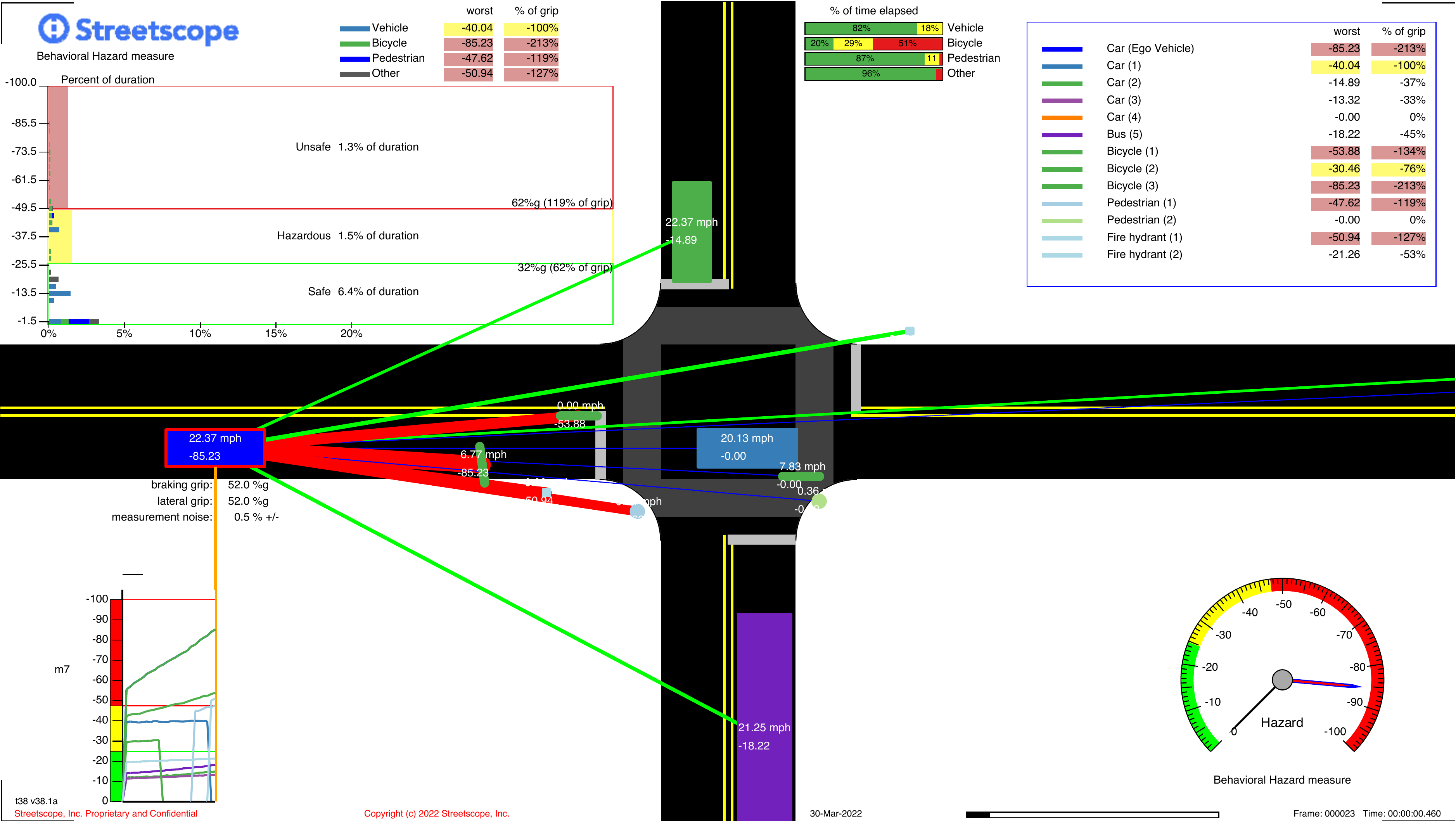}}}
  \vspace{8pt}
  \centerline{\fbox{\includegraphics*[height=\figheight]{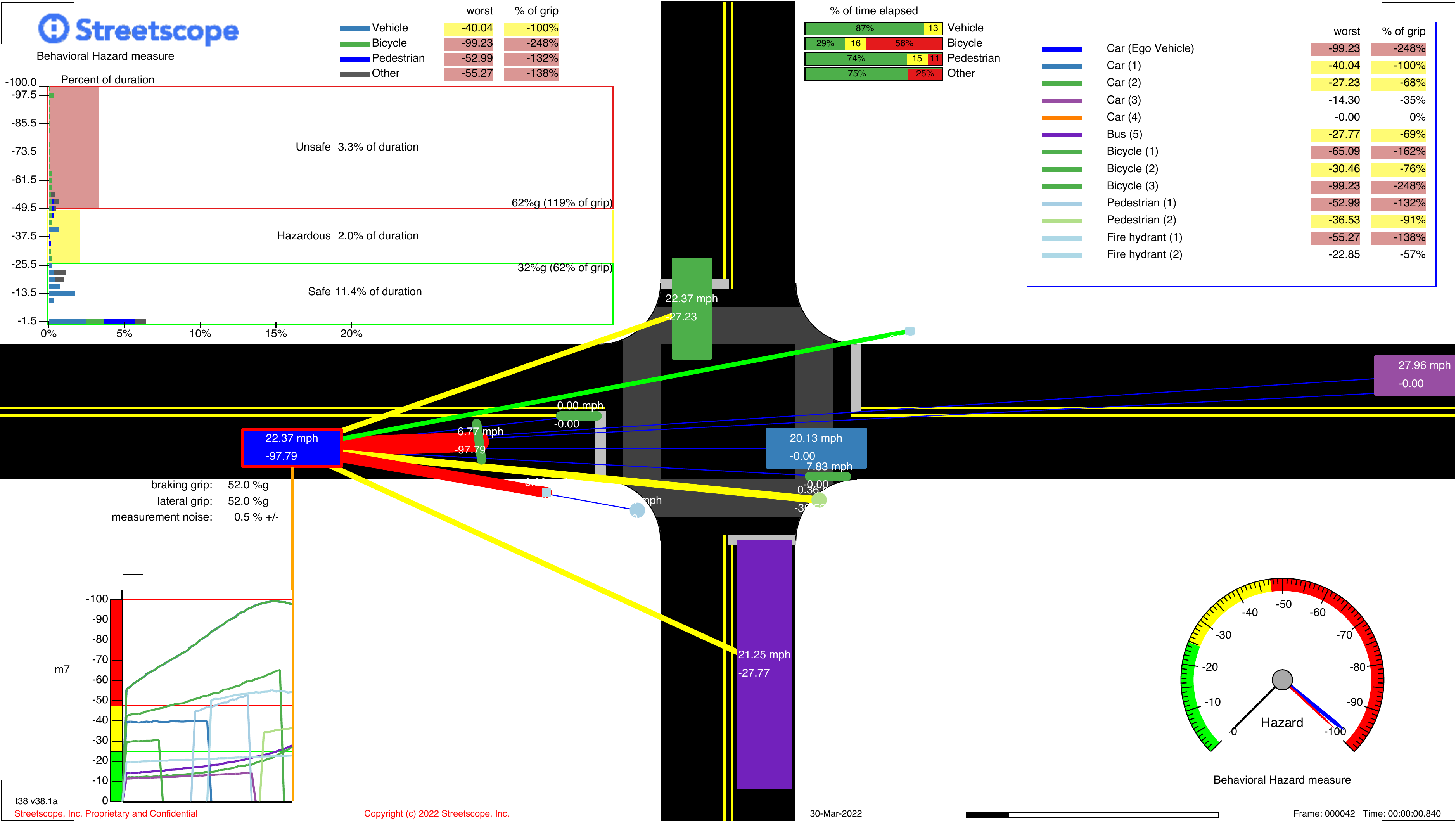}}}
  \vspace{8pt}
  \centerline{\fbox{\includegraphics*[height=\figheight]{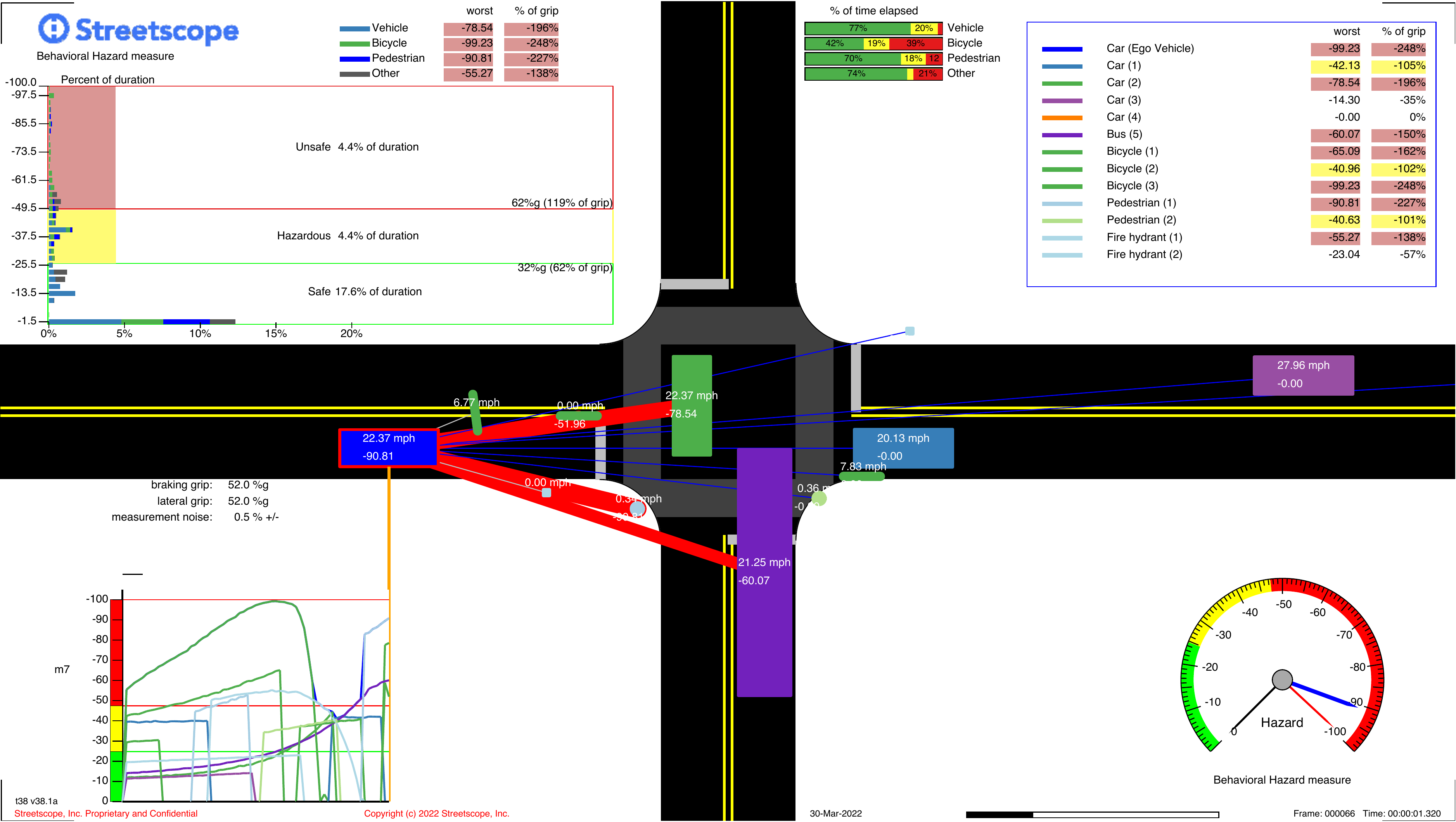}}}
  \caption{Three frames from a simulation of the collision hazard measure.}
  \label{fig:sim}
\end{figure}


\clearpage

A graph of all hazard values is being created in the bottom left corner of
each frame of the simulation.  The vertical axis is the value of the
SHM, and the horizontal axis corresponds to the
position of the center of the ego vehicle.
Each line, colored to match the color of the traffic object that generated
the collision hazard value, is displayed.
As can easily be seen, these values rise and fall as the separation distance
grows smaller and larger (respectively) and/or the relative speed grows
larger and smaller (respectively).

A histogram of the SHM values encountered up to this
frame is being created in the top left corner of the simulation frame,
along with various other representations and aggregations
of the SHM values.

\subsection{On-street Data}

\begin{sidewaysfigure}[ht]
  \includegraphics*[width=1.00\textwidth]{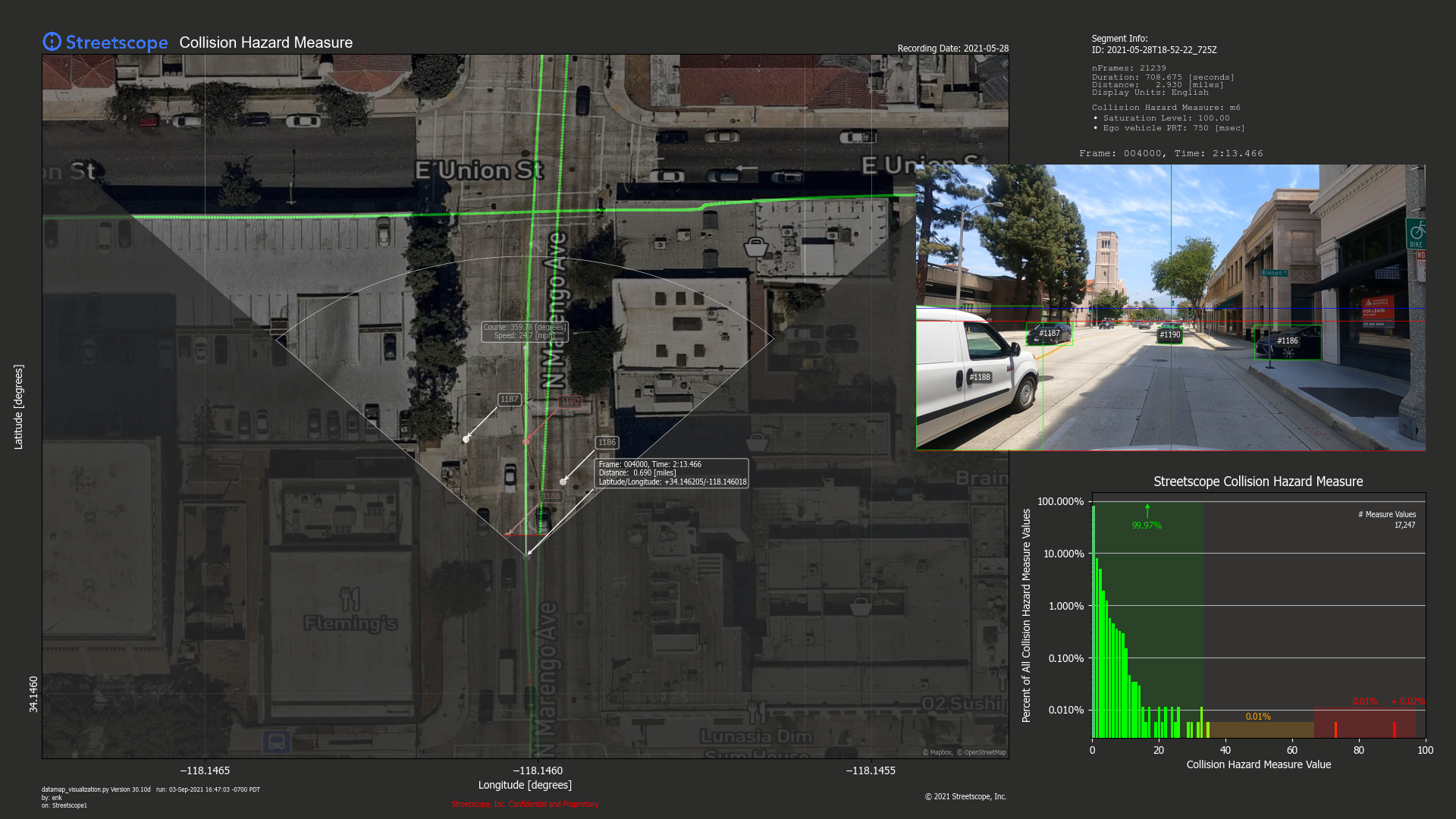}
  \caption{An example frame of a visualization of on-street data with
    the collision hazard measure values.}
  \label{fig:on-street}
\end{sidewaysfigure}

Over 2 years of data has been collected and analyzed from vehicles operating
on streets in traffic.
One frame from a visualization of analyzed data is included in
Figure~\ref{fig:on-street}, as an example.

SHM performed well with the complexities of vehicles in real urban
traffic, real sensors (with the attendant sensor noise,
object visibility, detection and tracking anomalies, and consequent
data processing challenges,
\textit{e.g.},
estimating position and speed of objects from monocular imagery)\@.

Future publications will examine on-street data in greater detail.

\clearpage
\stepcounter{ntab}
\renewcommand{\area}{Conclusions}
\renewcommand{\subarea}{}
\pagestyle{wicked}
\rightoddhead{\value{ntab}}{\area}{\subarea}{tabcolor07}
\leftevenhead{\value{ntab}}{\area}{\subarea}{tabcolor07}
\fancyhead[LO,RE]{\slshape{\area}} 
\fancyhead[CO,CE]{\slshape{\subarea}} 
\section{Conclusions}

Leading indicators for
measuring and assessing the safe operation of vehicles in traffic
are essential to the deployment of automated mobility.
Existing approaches, such as
TTC, RSS, ISM, IMU data, and disengagements, do not provide the information
required to provide a measurement of safety that will be useful
to regulators and the public.
The existing methods all require assumptions and/or predictions
about the behaviors of traffic objects, which will not, in general,
be correct, limiting the value of the information that they provide.

Lagging indicators, such as
historical collision statistics, do not provide timely information,
and require the ethically unacceptable occurrence of collisions,
property damage, injuries and deaths.

The Streetscope collision hazard measure (SHM)
has the essential characteristics
to provide a measurement of safety that will be useful
to regulators and the public.
It overcomes the limitations of existing measures,
provides an independent leading indication of safety,
and does not require assumptions nor predictions of behaviors.
The measure will also be directly useful for vehicle developers,
to independently monitor and assess engineering progress;
to fleet operators to monitor and improve ongoing operational safety;
to traffic managers to identify areas that generate high hazards and to
guide remediation and design efforts;
and to insurers to facilitate effective analysis of risk.

Many areas of application of the SHM include route selection
(whether for human-driven or automated vehicles),
to balance the competing objectives of speed, fuel efficiency, and safety.
Routes that include areas (such as intersections) that have a history
of many high-value near-misses would be penalized, and therefore avoided,
in the balanced selection.

\clearpage
\appendix
\section{Measurement Scales}
\label{app:measurement_scales}

\subsection{Properties}

Each scale of measurement satisfies one or more of the following
properties of measurement.~\cite{stattrekScales}
\begin{itemize}
\item \textbf{Identity}\@.
  Each value on the measurement scale has a unique meaning.
\item \textbf{Magnitude}\@.
  Values on the measurement scale have an ordered relationship
  to one another. That is, some values are larger and some are smaller.
\item \textbf{Equal intervals}\@.
  Scale units along the scale are equal to one another.
  This means, for example, that the difference between 1 and 2
  would be equal to the difference between 19 and 20.
\item \textbf{Zero}\@.
  A minimum value of zero.
  The scale has a true zero point, below which no values exist.
\end{itemize}

\subsection{Selected Scale Types}

\begin{quote}
\textbf{Ordinal Scale of Measurement}: The ordinal scale has the
property of both identity and magnitude. Each value on the ordinal
scale has a unique meaning, and it has an ordered (and therefore monotonic)
relationship to every other value on the scale.~\cite{stattrekScales}
\end{quote}

\begin{quote}
\textbf{Ratio Scale of Measurement}: The ratio scale of measurement
satisfies all four of the properties of measurement: identity,
magnitude, equal intervals, and a minimum value of zero.~\cite{stattrekScales}
\end{quote}

\subsection{Ratio Scales}
\label{app:ratio_scales}

\begin{quote}
\textit{Ratio scales} are those most commonly encountered in physics and
are possible only when there exist operations for determining all four
relations: equality [identity], rank-order [magnitude],
equality of intervals, and
equality of ratios [zero]\@.
\ldots
All types of statistical measures are applicable
to \textit{ratio scales} \ldots\@.~\cite{stevens1946measurement}
\end{quote}

\begin{quote}
[The \textit{ratio scale}] level of data measurement
allows the researcher to compare both
the differences and the relative magnitude of numbers. Some examples
of ratio scales include length, weight, time,
\textit{etc}\@.~\cite{formplusScales}
\end{quote}

%

\clearpage
\renewcommand{\area}{References}
\renewcommand{\subarea}{}
\pagestyle{wicked}
\addtocounter{ntab}{2}
\rightoddhead{\value{ntab}}{\area}{\subarea}{tabcolor99}
\leftevenhead{\value{ntab}}{\area}{\subarea}{tabcolor99}
\fancyhead[LO,RE]{\slshape{\area}} 
\fancyhead[CO,CE]{\slshape{\subarea}} 
{\footnotesize\nocite{vt2022white,ieee2846}
\bibliographystyle{IEEEtran}
\bibliography{streetscope, eka_patents}
}
\label{endlabel}%
\ifthenelse{\isodd{\pageref{endlabel}}}{\clearpage\blankpage}{}
\end{document}